\documentclass{article}
\usepackage{microtype}
\usepackage{graphicx}
\usepackage{booktabs}
\usepackage{hyperref}

\usepackage[accepted]{icml2025}
\usepackage{amssymb}
\usepackage{mathtools}
\usepackage{amsthm}
\usepackage[capitalize,noabbrev]{cleveref}
\theoremstyle{plain}

\theoremstyle{definition}

\theoremstyle{remark}

\usepackage[textsize=tiny]{todonotes}

\usepackage{caption}
\usepackage{subcaption}
\usepackage{multirow}
\usepackage{listings}
\usepackage{enumitem}
\usepackage{soul}
\definecolor{myyellow}{rgb}{1, 0.972549, 0.584313}
\definecolor{mygreen}{rgb}{0.772549, 0.945098, 0.756862}
\definecolor{myblue}{rgb}{0.811764, 0.866666, 0.996078}
\definecolor{myred}{rgb}{0.984313, 0.749019, 0.737254}
\definecolor{darkgreen}{rgb}{0, 0.5001960, 0}
\definecolor{darkred}{rgb}{0.8, 0, 0}

\icmltitlerunning{A Survey on Prompt Tuning}

\begin{document}

\twocolumn[
\icmltitle{A Survey on Prompt Tuning}

\begin{icmlauthorlist}
\icmlauthor{Zongqian Li}{1}
\icmlauthor{Yixuan Su}{1}
\icmlauthor{Nigel Collier}{1}
\end{icmlauthorlist}

\icmlaffiliation{1}{University of Cambridge}

\icmlcorrespondingauthor{Zongqian Li, Nigel Collier}{zl510@cam.ac.uk, nhc30@cam.ac.uk}

\vskip 0.3in
]

\printAffiliationsAndNotice{}

\begin{abstract}
This survey reviews prompt tuning, a parameter-efficient approach for adapting language models by prepending trainable continuous vectors while keeping the model frozen. We classify existing approaches into two categories: direct prompt learning and transfer learning. Direct prompt learning methods include: general optimization approaches, encoder-based methods, decomposition strategies, and mixture-of-experts frameworks. Transfer learning methods consist of: general transfer approaches, encoder-based methods, and decomposition strategies. For each method, we analyze method designs, innovations, insights, advantages, and disadvantages, with illustrative visualizations comparing different frameworks. We identify challenges in computational efficiency and training stability, and discuss future directions in improving training robustness and broadening application scope.\footnote{\url{https://github.com/ZongqianLi/Prompt-Tuning-Survey}}
\end{abstract}

\section{Introduction}

Large Language Models (LLMs) have achieved success across various Natural Language Processing (NLP) tasks, yet their adaptation to downstream applications faces challenges due to the computational and storage costs of full model fine-tuning \citep{zhao2024surveylargelanguagemodels}. Prompt tuning has emerged as a promising parameter-efficient fine-tuning (PEFT) approach that offers several advantages: (1) parameter efficiency through updating only a small group of continuous vectors while keeping the pretrained language model frozen; (2) modular adaptation through task-specific prompts that preserve the original model and parameters, enabling efficient deployment; (3) framework flexibility supporting various knowledge transfer and composition mechanisms, facilitating multi-task learning and domain adaptation. \citep{PromptTuning} 

The growing significance of prompt tuning is evidenced by its dual influence: for the academic community, it intersects with various areas including model adaptation, knowledge transfer \citep{44873}, continual learning \cite{10444954}, and model interpretability \citep{10444954}; for commercial applications, it enables economical model customization, reduces deployment overhead, and accelerates application development \citep{li2025reasongraphvisualisationreasoningpaths, D4DD00307A}. This widespread adoption across both domains necessitates a survey to understand its underlying mechanisms, framework designs, innovations, and development trends.

\begin{figure*}[th]
    \centering
    \includegraphics[width=0.99\textwidth]{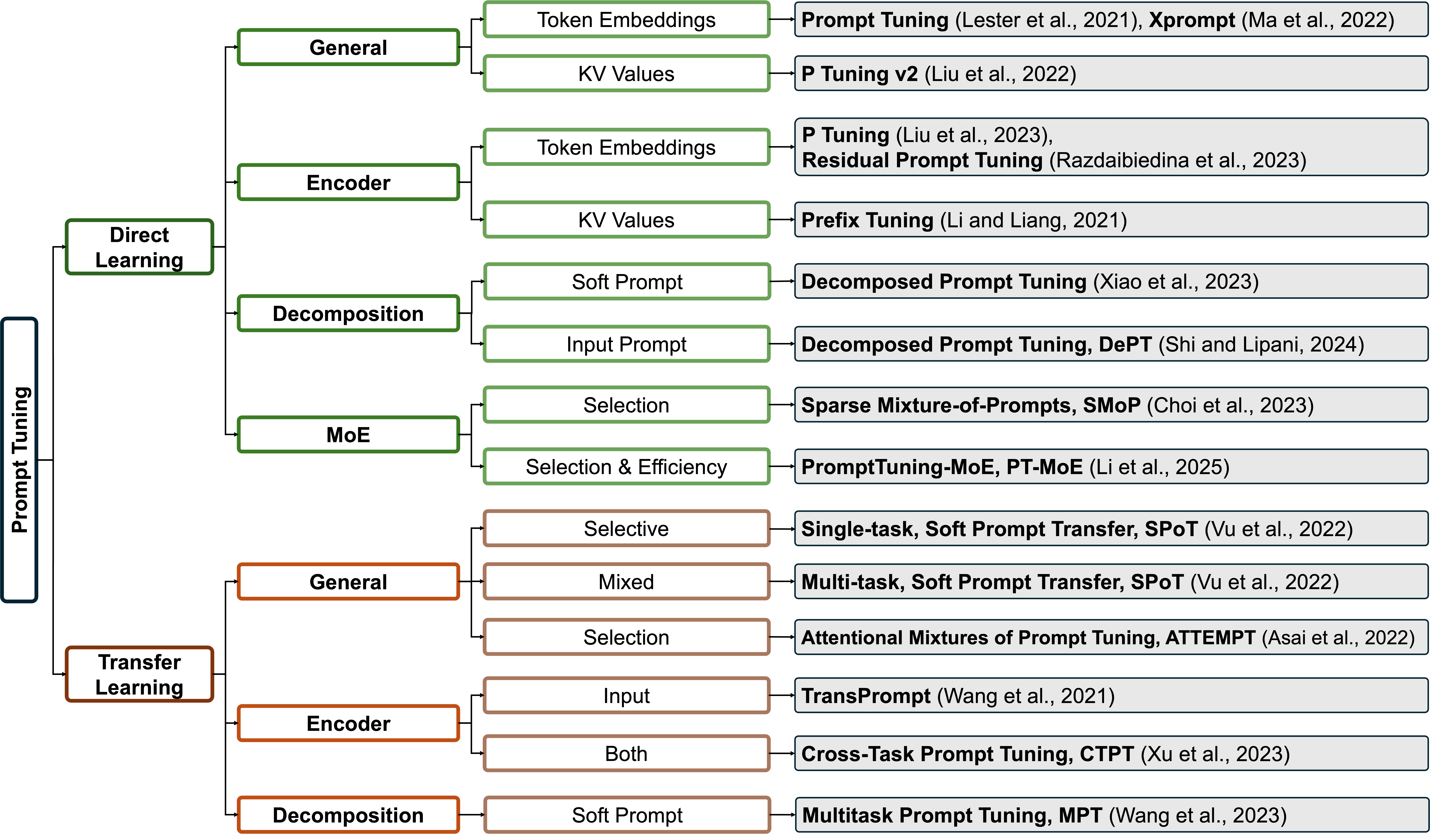}
    \caption{Hierarchical overview of prompt tuning methods including direct learning and transfer learning.}
    \label{overview}
\end{figure*}

Unlike existing surveys or papers that broadly review PEFT \citep{han2024parameterefficient} or general prompting methods \citep{schulhoff2024}, this survey focuses on prompt tuning approaches and makes three primary contributions:

\vspace{-5pt}

\begin{itemize}[left=0pt, itemsep=0pt, parsep=0pt]
    \item We present a comprehensive categorization of prompt tuning approaches, categorizing existing methods into two main branches as shown in Figure \ref{overview}: direct prompt learning and transfer learning. Direct prompt learning encompasses methods that perform single-stage training on target tasks. Transfer learning methods, on the other hand, focus on leveraging knowledge from source tasks to improve performance on target tasks.
    \item For each method, we provide an in-depth analysis of its framework, principles, innovations, key insights, and relative advantages and limitations;
    \item We provide illustrative visualizations to compare different methods.
    \item We identify challenges in current prompt tuning methods and propose future directions.
\end{itemize}

\vspace{-5pt}

This survey is organized as follows: Section \ref{Preliminary} introduces the preliminaries of PEFT, soft prompts, and transfer learning. Section \ref{DirectPromptLearning} and Section \ref{TransferLearning} present a review of direct prompt learning and transfer learning approaches, respectively. Section \ref{ChallengesandFutureWork} discusses current challenges and future directions, followed by conclusions in Section \ref{Conclusions}.

\section{Preliminary}
\label{Preliminary}

\textbf{PEFT} aims to adapt pre-trained language models with minimal trainable parameters. These methods can be broadly categorized into: (1) addition-based methods that introduce new parameters inserted between transformer layers \citep{pmlr-v97-houlsby19a}; (2) selection-based methods that update specific types of parameters while freezing other components \citep{ben-zaken-etal-2022-bitfit}; (3) reparameterization-based methods that decompose weight updates into low-rank matrices to reduce trainable parameters \citep{hu2022lora}; (4) prompt-based methods that prepend trainable vectors to the input sequence \citep{PromptTuning}. These approaches modify a small amount of model parameters while achieving comparable performance to full fine-tuning. Parameter-efficient methods reduce computational resources and memory requirements during training and deployment, making large-scale model adaptation more accessible \citep{wan2024efficient}. However, they often require careful hyperparameter tuning and may exhibit performance decrease on complex tasks or small-scale pre-trained models.

\textbf{Soft prompts} are continuous vectors in the embedding space, different from discrete prompts that consist of natural language tokens (with trained and fixed word embeddings) \citep{zhao-etal-2023-spc}. Formally, given an input sequence $\mathbf{x} = [x_1, ..., x_n]$, soft prompts $\mathbf{P} \in \mathbb{R}^{m \times d}$ are prepended to create a lengthened input sequence $[\mathbf{P}; E(\mathbf{x})]$, where $E(\cdot)$ denotes the embedding layer, $m$ is the length of the soft prompt, and $d$ is the dimension of the embedding space. Soft prompts have two key characteristics: (1) they are unrestricted by vocabulary constraints, enabling optimization in a continuous space and thus encoding task-specific information more flexibly than discrete prompts; (2) they maintain the same dimensionality as word embeddings, ensuring compatibility with transformer. These prompts can be initialized either randomly or from existing word embeddings, leading to different optimization processes.

\begin{figure*}[th!]
    \centering
    \includegraphics[width=0.76\textwidth]{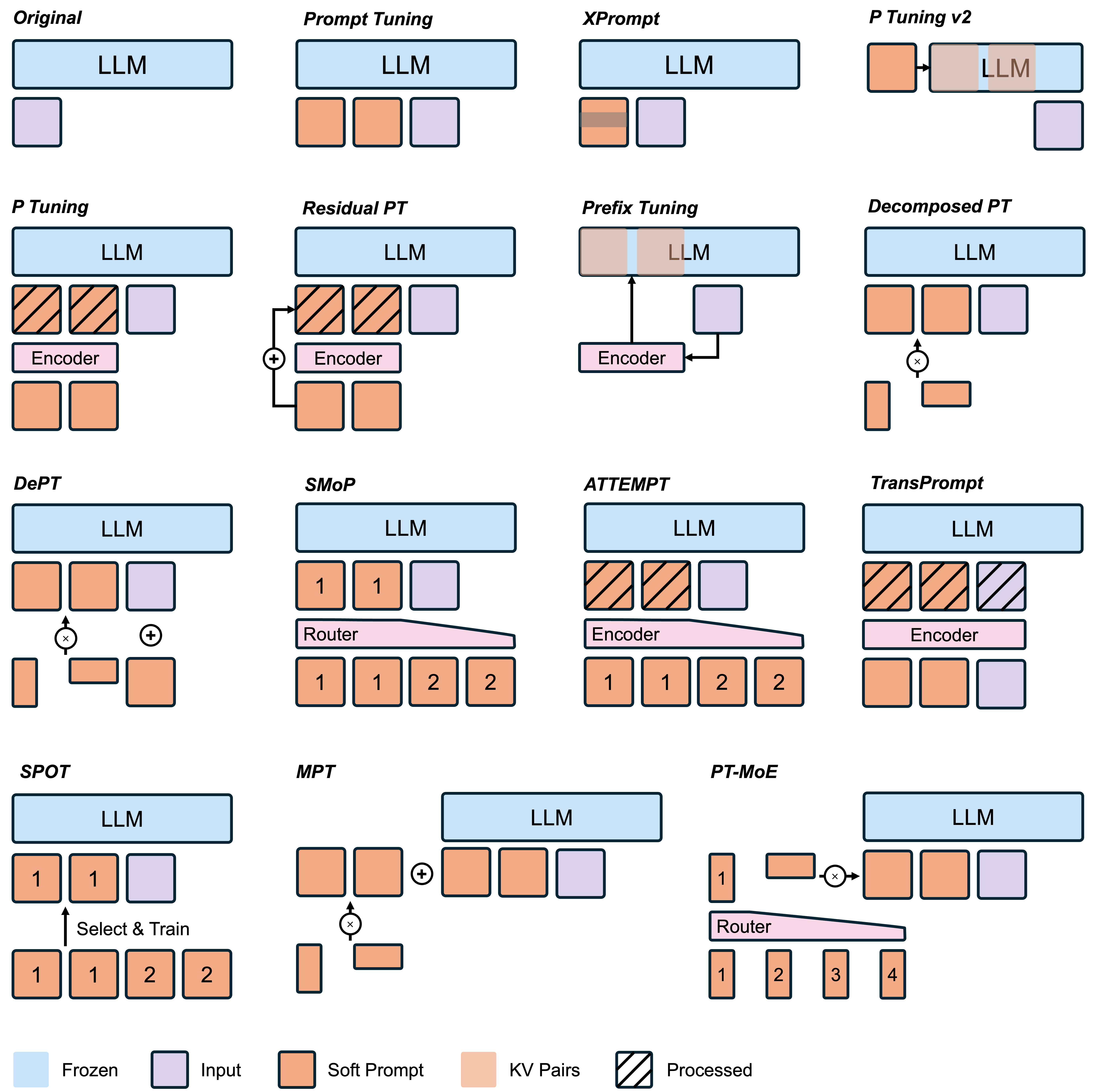}
    \caption{Illustration of different prompt tuning (PT) methods. Prompt Tuning directly prepends soft prompts to input. XPrompt applies pruning to soft prompts. P-Tuning v2 and Prefix Tuning incorporate prompts or KV pairs across all the layers of the language model. P-Tuning and Residual PT use encoders to process soft prompts. Decomposed PT and DePT leverage matrix decomposition strategies. SPoT leverages pre-trained prompts from source tasks, ATTEMPT or CTPT utilizes prompt mixing, TransPrompt uses encoders for task-specific and universal knowledge, while MPT decomposes prompts into shared and task-specific components.}
    \label{all-methods}
\end{figure*}

\textbf{Transfer learning} in prompt tuning leverages knowledge from source tasks to improve performance on target tasks \citep{gu-etal-2022-ppt}. Source tasks typically refer to large-scale datasets or tasks with abundant labeled data, while target tasks are usually low-resource scenarios where labeled data is limited \citep{9134370}. This framework comprises three essential components: (1) knowledge transfer strategies that address how to extract and encode knowledge from source tasks; (2) task adaptation mechanisms that determine how to utilize transferred knowledge for target tasks; (3) cross-domain methods that bridge gaps between source and target domains, such as handling different label spaces. The effectiveness of transfer learning in prompt tuning depends on several variables: task similarity between source and target domains, quality of source task data, and the robustness of the transfer mechanism. Studies have proved that well-designed transfer approaches can improve prompt tuning performance, particularly in few-shot scenarios where target task data is limited.

\section{Direct Prompt Learning}
\label{DirectPromptLearning}

Direct prompt learning methods typically involve a single training stage where soft prompts are optimized on target tasks while keeping the pretrained language model parameters frozen. These approaches can be categorized into four main branches: General methods that directly tune prompt embeddings or KV pairs like Prompt Tuning \citep{PromptTuning}, XPrompt \citep{XPrompt}, and P-Tuning v2 \citep{PTuningv2}; Encoder-based methods that incorporate additional layers to process prompts such as P-Tuning \citep{PTuning}, Residual Prompt Tuning (RPT) \citep{ResidualPromptTuning}, and Prefix Tuning \citep{PrefixTuning}; Decomposition-based methods that decompose prompt embeddings including Decomposed Prompt Tuning (DPT) \citep{DecomposedPromptTuning} and DePT \citep{DePT}; and MoE approaches that leverage multiple soft prompts like Sparse Mixture-of-Prompts (SMoP) \citep{SMoP}. Direct prompt learning methods are illustrated in Figure \ref{all-methods}.

\textbf{Prompt Tuning} freezes all the parameters of the original pretrained language model and prepends a list of trainable tokens to the input tokens \citep{PromptTuning}. These trainable tokens are initialized by random values or specific words embeddings, and are optimized to adapt the language model for downstream tasks. Prompt Tuning is the first method that finetunes the input tokens, which proves that training a small amount of parameters can reach similar performance with full finetuning. The larger the language model is, the more effective the method is. For T5-XXL with 11B parameters, prompt tuning matches the performance of full tuning within a couple percentage points on most tasks in SuperGLUE. Less additional parameters save the time for training the language model and the memory to store the trained models. However, the training process for prompt tuning is difficult to converge and is dependent on the initialization of the trainable tokens and learning rate. Sufficient training data are needed as well.

\textbf{XPrompt} is an efficient prompt tuning method that utilizes pruning to soft prompt tokens \citep{XPrompt}. This approach consists of three steps: (1) vanilla prompt tuning on the target tasks; (2) hierarchical pruning, which first removes negative soft prompt tokens at token-level and then uses fine-grained pruning at piece-level; (3) weight rewinding to retrain the identified positive soft prompt tokens. XPrompt is based on lottery ticket hypothesis. This theory proposes that an over-parameterized layer contains smaller sublayers that can achieve similar or better performance when trained independently from the same initialization. The key insight of XPrompt is the first application of lottery ticket hypothesis to prompt tuning, revealing that not all soft prompt tokens contribute equally to model performance and some tokens may even have negative influences. This observation leads to the innovation of piece-level pruning within each token. XPrompt improves model performance across different model scales, achieves higher parameter efficiency, and works well in both data-rich and few-shot scenarios. However, one question left is how to find the optimal compression ratio without trial training.

\textbf{P-Tuning v2} improves prompt tuning by introducing deep prompts (soft prompts) across all layers of pre-trained language models while keeping the original parameters frozen \citep{PTuningv2}. Different from traditional prompt tuning that only adds trainable continuous prompts to the input layer, P-Tuning v2 incorporates prefix trainable tokens as soft prompts in every transformer layer. This deep prompt design eliminates the need for verbalizers and language model heads in favor of a simple classification head. The key insights of P-Tuning v2 include: (1) prompt depth influences model performance and adding soft prompts to deeper layers can match the performance of full-layer prompting; (2) optimal prompt length varies by task complexity with simple classification tasks preferring shorter soft prompts while sequence labeling tasks benefit from longer ones; (3) reparameterization influences are task-dependent rather than universally beneficial. P-Tuning v2 achieves comparable performance to full fine-tuning across different model scales (330M-10B) and diverse NLP tasks, demonstrating strong parameter efficiency. However, it requires task-specific hyperparameter tuning and may underperform full fine-tuning on certain sequence labeling tasks, particularly with smaller models.

\textbf{P-Tuning} introduces a hybrid method that uses trainable continuous prompt embeddings while incorporating discrete text tokens at specific positions, freezing the pretrained model parameters \citep{PTuning}. Different from conventional prompt tuning, the continuous prompt embeddings in P-Tuning can be interrupted by discrete word tokens and a prompt encoder (LSTM or MLP) is employed to learn dependencies between continuous embeddings. There are several insights for P-Tuning: (1) continuous and discrete prompts complement each other, with continuous prompts providing optimization flexibility while discrete prompts maintain fiengrained information; (2) prompt placement influences task performance, with better results achieved when continuous prompts do not influence the completeness of sentences; (3) the prompt encoder influences training stability with LSTM and MLP generally outperforming direct embedding optimization. P-tuning improves both model performance and training stability while reducing the variance between different prompts. However, this approach introduces additional trainable parameters for the prompt encoder, requires task-specific tuning of prompts, and shows decreasing returns on large-scale tasks where traditional fine-tuning remains competitive.

\textbf{RPT} introduces a novel reparameterization approach that improves continuous prompt embeddings through a residual design while maintaining the frozen pretrained model \citep{ResidualPromptTuning}. The continuous embeddings are transformed through a residual layer before being prepended to inputs, where the MLP consists of down-projection and up-projection layers, followed by layer normalization and a skip connection to preserve the original embedding information. Several key insights emerge from this method: (1) the residual connection facilitates optimization by providing a direct path for preserving original embedding information while allowing the MLP to focus on learning useful transformations, leading to faster convergence and better performance; (2) the shared reparameterization layers exhibits superior performance in low-resource scenarios; (3) the compression strategy in the MLP design proves crucial for performance, with increased dimensionality leading to consistent improvements until saturation. RPT demonstrates improvements in performance stability and hyperparameter robustness. However, this approach still exhibits limitations: it introduces additional parameters through the reparameterization layer and maintains a performance gap with full fine-tuning.

\textbf{Prefix-Tuning} introduces a parameter-efficient approach that prepends trainable continuous key-value vectors at each transformer layer without modifying the pretrained model parameters \citep{PrefixTuning}. Unlike prompt tuning that only optimizes embeddings at the input layer, Prefix-Tuning directly optimizes prefix keys and values across all transformer layers. Specifically, for each transformer layer, the prefix key-value pairs are generated through a reparametrization layer: a smaller trainable matrix is transformed through an MLP to produce the layer-specific key-value vectors. Several key insights emerge from the empirical analysis: (1) direct optimization of prefix parameters leads to instability, necessitating the reparameterization strategy for robust training; (2) this method demonstrates superior performance in low-resource scenarios, surpassing full fine-tuning; (3) preserving pretrained parameters benefits cross-domain generalization. While Prefix-Tuning achieves parameter efficiency and enables efficient multi-task batching, it exhibits certain limitations: performance gap on large-scale datasets compared to full fine-tuning, and dependence to prefix length and initialization strategies.

\textbf{DPT} reparameterizes the soft prompts in prompt tuning through low-rank matrix decomposition \cite{DecomposedPromptTuning}. Prompt tuning initializes soft prompts either randomly or from vocabulary embeddings, however, DPT decomposes the original prompt matrix $\mathbf{P}_{\text{emb}} \in \mathbb{R}^{e \times c}$ into the product of two smaller matrices $\mathbf{A} \in \mathbb{R}^{e \times b}$ and $\mathbf{B} \in \mathbb{R}^{b \times c}$, where $b$ denotes a small bottleneck dimension. Several key insights emerge: (1) soft prompts naturally tend to converge towards low-rank forms during training, suggesting inherent dimensionality redundancy; (2) direct parameterization of this low-rank property through matrix decomposition offers a more effective way to learn prompt embeddings; (3) the bottleneck dimension influences the complexity of prompt embeddings. While DPT achieves parameter efficiency and performance improvements and demonstrates robust performance in few-shot scenarios, it exhibits certain limitations: inherited slow convergence from prompt tuning, dependence to bottleneck dimension selection, and unverified effectiveness on generation tasks beyond natural language understanding.

\textbf{DePT} improves the prompt tuning method through a decomposition strategy as well \citep{DePT}. Unlike DPT which decomposes the prompt matrix through low-rank matrix decomposition, DePT proposes a hybrid approach that decomposes the original prompt matrix $\mathbf{P} \in \mathbb{R}^{l \times d}$ into a shorter soft prompt $\mathbf{P}_s \in \mathbb{R}^{m \times d}$ and a pair of low-rank matrices $\mathbf{A} \in \mathbb{R}^{s \times r}$ and $\mathbf{B} \in \mathbb{R}^{r \times d}$, where $r$ denotes the rank dimension and $s$ is the maximum sequence length. The low-rank matrices are used to update the frozen word embeddings $\mathbf{W}$ through element-wise addition ($\mathbf{W'} = \mathbf{W} + \mathbf{BA}$), while the shorter soft prompt is prepended to the input sequence. The investigation reveals several key insights: (1) the combination of low-rank updates and soft prompts maintains competitive performance while enabling efficient parameter utilization due to reduced sequence length; (2) employing dual learning rates ($\alpha_1$ for soft prompts and $\alpha_2$ for low-rank matrices) is crucial for optimization convergence; (3) updating word embeddings through low-rank matrices improves the adaptability of inputs. DePT demonstrates advantages: it achieves performance improvements, reduces computational costs, shows increasing efficiency gains with larger models, and demonstrates robust integration with parameter-efficient transfer learning frameworks particularly in few-shot learning scenarios. However, it exhibits limitations: increased hyperparameter complexity due to dual learning rates, dependency on maximum sequence length for parameter allocation, and performance gaps compared to full fine-tuning in specific tasks.

\textbf{SMoP} reformulates the prompt tuning through mixture-of-prompts mechanism \citep{SMoP}. While standard prompt tuning employs a single continuous soft prompt $\mathbf{P}_{\theta} \in \mathbb{R}^{l \times e}$ of length $l$, SMoP parameterizes multiple shorter prompts ${\mathbf{P}_{\theta_j}} \in \mathbb{R}^{l' \times e}$ where $l' \ll l$, coupled with a gating mechanism that dynamically routes each input embedding to the most suitable soft prompt. The investigation reveals several fundamental insights: (1) the performance gains from increasing prompt length become marginal beyond a large number of tokens, indicating inherent redundancy in long prompts; (2) the number of prompts $k$ exhibits an optimal range determined by data availability, with excessive prompts leading to decreased performance due to insufficient training data per prompt; (3) maintaining balanced prompt utilization is crucial for optimal performance. While SMoP demonstrates efficiency improvements and achieves performance gains compared to standard prompt tuning, it faces limitations: the gating mechanism introduces additional parameters, and its effectiveness depends on sufficient training data for each prompt. 

\textbf{PT-MoE} integrates matrix decomposition with mixture-of-experts routing to improve parameter efficiency and task performance in prompt tuning \citep{li2025ptmoeefficientfinetuningframework}. Unlike conventional prompt tuning that uses static prompts, PT-MoE decomposes each soft prompt $P_i$ into an input-specific matrix $A_i \in \mathbb{R}^{T \times R}$ and a shared matrix $B \in \mathbb{R}^{R \times H}$, where $T$, $R$, and $H$ denote the prompt length, low-rank dimension, and hidden dimension respectively. The final prompt is computed as $P = \sum_{i=1}^N w_i A_i B$ using decomposed matrices $A_i$ and $B$, weights $w$, and the number of experts $N$. The framework uses a linear router with Gaussian noise during training, followed by hard selection via straight-through estimation. Several key insights emerge from this approach: (1) matrix decomposition enables efficient parameter sharing across experts while maintaining task-specific adaptability; (2) MoE routing provides dynamic adaptation based on input characteristics, yielding complementary benefits when combined with matrix decomposition; (3) the integration demonstrates cross-task consistency, achieving state-of-the-art performance in both question answering and mathematical reasoning tasks. While PT-MoE achieves superior performance with 25\% fewer parameters than LoRA and outperforms other prompt tuning methods, it has limitations such as the need for an additional router.

\section{Transfer Learning}
\label{TransferLearning}

Transfer learning approaches in prompt tuning leverage knowledge from source tasks to improve performance on target tasks through prompt transfer or adaptation. These methods can be classified into three main groups: General methods that directly transfer or mix prompt knowledge like Soft Prompt Transfer (SPoT) \citep{SPoT} and Attentional Mixtures of Prompt Tuning (ATTEMPT) \citep{ATTEMPT}; Encoder-based methods that utilize additional layers to facilitate knowledge transfer such as TransPrompt \citep{TransPrompt} and Cross-Task Prompt Tuning (CTPT) \citep{CTPT}; and Decomposition-based methods that decompose prompts into shared and task-specific components like Multitask Prompt Tuning (MPT) \citep{MPT}. Transfer learning methods for prompt tuning are illustrated in Figure \ref{all-methods}.

\textbf{SPOT} introduces transfer learning into prompt tuning through two transfer strategies, both followed by continued training on target tasks \citep{SPoT}. The first strategy, termed generic transfer, trains a general soft prompt on multiple source tasks and uses it to initialize prompts for all target tasks, which are then further tuned on the respective tasks. The second strategy, termed targeted transfer, adopts a retrieval-based approach: it first trains task-specific prompts for each source task independently, preserving both early checkpoints (for task similarity comparison) and final checkpoints (for transfer initialization). For a new target task, it measures the similarity between this task and source tasks using the early checkpoints of the prompts for source tasks, selects the most similar source task, and initializes the target prompt using the final checkpoint of that task, followed by continued prompt tuning on the target task. Several key insights emerge from SPoT: (1) prompt transfer can effectively improve performance; (2) task similarity strongly correlates with transfer effectiveness, with high-quality source tasks being either large-scale datasets, complex reasoning tasks, or tasks similar to the target; (3) early checkpoints of prompts serve as better task embeddings than final checkpoints for measuring task similarity, suggesting the importance of capturing task-specific knowledge in the early training stage. While SPOT matches or outperforms full model fine-tuning across all model sizes while only tuning a small number of parameters, it exhibits certain limitations: high dependence to source task selection, incomplete capture of transfer characteristics through task embeddings, and additional computational overhead for computing task embeddings.

\textbf{ATTEMPT} introduces an attention-based prompt mixing mechanism for efficient multi-task knowledge transfer in prompt tuning \citep{ATTEMPT}. The method first pre-trains a group of source prompts $[\mathbf{P}_1,...,\mathbf{P}_t]$ on large-scale source tasks. For each target task input, ATTEMPT computes attention $a_j$ between the input embedding and each prompt through an attention component $G$, which consists of down-projection, non-linear transformation, and up-projection layers. The input-level prompt is then generated as a weighted combination: $\mathbf{P}_{\text{input}} = \mathbf{P}_{\text{target}} + \sum_{j=1}^{t+1} a_j\mathbf{P}_j$, where $\mathbf{P}_{\text{target}}$ is a learnable target-task prompt. During training, only $\mathbf{P}_{\text{target}}$ and $G$ are updated while keeping the language model frozen. Several key insights emerge from their investigation: (1) the effectiveness of prompt tuning exhibits strong scaling behavior and performance improves with larger base models despite poor results on smaller ones; (2) various source prompts contribute differently to target tasks, necessitating dynamic input-level adaptation; (3) knowledge transfer exists between ostensibly different tasks, suggesting broader applicability of prompt reuse. While ATTEMPT achieves parameter efficiency and competitive performance, it faces certain limitations: increased memory overhead due to extended input sequences, decreased performance with small base models, and dependency on pre-trained source prompts.

\textbf{TransPrompt} reformulates prompt tuning as a cross-task knowledge transfer framework through a dual-encoder design \citep{TransPrompt}. While normal prompt tuning methods may utilize a single prompt encoder, TransPrompt simultaneously employs two parallel encoders: a task-specific encoder and a universal encoder. Specifically, given an input $x$, both encoders process the same templated sequence. The task-specific encoder uses task-dependent prompt templates, while the universal encoder employs shared prompt templates. Both encoders are implemented as BiLSTM followed by MLPs. The outputs from both encoders are combined through average pooling before being fed into the language model for final prediction. Several key insights emerge from the investigation: (1) the dual-encoder design enables simultaneous capture of task-specific and universal knowledge; (2) normalization mechanisms, specifically prototype-based and entropy-based techniques, are crucial for learning task-agnostic embeddings; (3) the universal prompt template effectively captures cross-task commonalities, facilitating generalization to unseen tasks. While TransPrompt demonstrates consistent performance improvements and exhibits strong generalization capabilities, it faces certain limitations: complex training procedure; computational overhead from the double encoder design, dependency on task similarity for effective transfer, and dependence to hyperparameters, particularly requiring careful learning rate tuning for optimal convergence.

\textbf{CTPT} introduces a cross-task prompt tuning framework comprising three key components: Task-specific Prompt Tuning (TSPT), Cross-Task Prompt Learning (CTPL), and Cross-Task Prompt Observation (CTPO) \citep{CTPT}. The method leverages both task-specific prompts $\mathbf{P}_{i}$ trained on few-shot examples of the target task and a group of source prompts ${\mathbf{P}_{j}}$ from pre-trained related tasks. Specifically, TSPT learns task-specific knowledge independently, while CTPL captures cross-task knowledge through multi-head attention where $\mathbf{P}_{i}$ serves as the query while ${\mathbf{P}_{j}}$ provide keys and values, followed by a residual connection: $\mathbf{P}_{\text{final}} = \mathbf{P}_{i} + \sum_{j}\text{softmax}(\frac{\mathbf{W}_Q\mathbf{P}_{i}\mathbf{W}_K^T\mathbf{P}_{j}}{\sqrt{d}})\mathbf{W}_V\mathbf{P}_{j}$
. CTPO then filters redundant information through a gate-like mechanism. The attention parameters $\mathbf{W} = [\mathbf{W}_Q; \mathbf{W}_K; \mathbf{W}_V]$ are obtained through a dimension reduction strategy: mapping from a low-dimensional vector $\mathbf{z}$ to the parameter space via a fixed matrix $\mathbf{A}$, $\mathbf{W} = \mathbf{A}\mathbf{z}$. The entire framework is optimized through derivative-free optimization (CMA-ES) in low-dimensional space. Several key insights emerge from the investigation: (1) emotional knowledge is inherently transferable across conversation datasets despite varying label nomenclatures; (2) performance improvement exhibits a linear correlation with the number of source tasks, indicating scalable knowledge transfer; (3) the intrinsic dimensionality of prompt parameters is substantially lower than their apparent dimensions. While CTPT demonstrates computational efficiency and robust zero-shot transfer capabilities, it exhibits certain limitations: requiring multiple forward passes for optimization, slower convergence compared to gradient-based methods, and decreased performance when source and target tasks differ significantly.

\textbf{MPT} reformulates prompt tuning through a decomposition of soft prompts into shared and task-specific components via knowledge distillation \citep{MPT}. While conventional prompt tuning learns independent prompts for each task, MPT parameterizes the task prompt $\mathbf{\hat{P}}_k$ as:
$\mathbf{\hat{P}}_k = \mathbf{P}^* \circ (\mathbf{u}_k \otimes \mathbf{v}_k^T)$
where $\mathbf{P}^* \in \mathbb{R}^{l \times d}$ is the shared prompt matrix, $\mathbf{u}_k \in \mathbb{R}^l$ and $\mathbf{v}_k \in \mathbb{R}^d$ form the rank-one task-specific matrix through outer product, $\circ$ denotes the Hadamard (element-wise) product, $l$ denotes prompt length and $d$ denotes embedding dimension. The framework employs a distillation objective combining KL-divergence for logit matching and MSE loss for hidden state to transfer knowledge from independently trained teacher prompts. During source training, MPT adopts a stochastic task sampling strategy that dynamically varies the number of tasks per batch. For target adaptation, it implements a dual learning rate scheme where task-shared and task-specific components are updated with distinct learning rates. Several key insights emerge from this investigation: (1) cross-task knowledge can be effectively compressed into a single shared prompt while preserving task-specific adaptability; (2) the decomposition of prompts into shared and task-specific components enables efficient knowledge transfer across tasks; (3) knowledge distillation from independently trained prompts facilitates learning better decomposable embeddings. While MPT demonstrates parameter efficiency and achieves competitive performance including few-shot scenarios, it exhibits certain limitations: complex training process, computational overhead from teacher model training and distillation process, dependence to prompt length selection, and suboptimal performance on complex tasks compared to full fine-tuning.

\section{Challenges and Future Work}
\label{ChallengesandFutureWork}

\subsection{Current Challenges}

\textbf{Computational efficiency}. The addition of soft prompts extends input sequence length, increasing memory consumption and computational costs. This overhead becomes substantial when processing long sequences or storing multiple task-specific prompts \citep{10.1145/3530811}.

\textbf{Training instability}. Prompt tuning demonstrates high dependence to hyperparameter choices, particularly learning rates, leading to convergence difficulties and inconsistent performance across different initializations \citep{gu-etal-2022-ppt}. This necessitates hyperparameter tuning to achieve stable and optimal results.

\textbf{Prompt initialization}. Initialization strategies for soft prompts, whether random or embedding-based, influence model performance and convergence, yet current approaches can yield suboptimal results \citep{liu2022fewshot}.

\textbf{Model scale dependency}. The effectiveness of prompt tuning correlates with model scale, performing well on large models but decreasing on smaller ones, limiting its applicability in resource-constrained settings \citep{PromptTuning}.

\textbf{Explainability}. The semantic meaning of learned prompts and their interaction mechanisms with pretrained models remain poorly understood, limiting their improvements.

\subsection{Future Directions}

\textbf{Training stability}. Future work should focus on developing more robust optimization strategies and adaptive learning techniques \citep{Liu2020On} that reduce hyperparameter dependence. This includes exploring adaptive prompt designs and innovative initialization strategies that ensure stable convergence.

\textbf{Scope extension}. Prompt tuning can extend to diverse-input tasks \citep{YAO202430}, multi-output prediction problems, and specific domains.

\textbf{Transfer learning}. Future work should explore meta-learning approaches \citep{nichol2018} for prompt adaptation, hierarchical prompt designs, and methods for handling different label spaces across tasks.

\textbf{Interpretability}. Key directions include analyzing optimization paths, developing explainability methods, and understanding connections with other efficient finetuning approaches \citep{han2024parameterefficient}.

\textbf{Parameter efficiency}. Key areas include developing lightweight prompt embeddings \citep{500xcompressor, li-etal-2025-prompt}, exploring dynamic generation methods, and investigating shared prompt designs.

\section{Conclusions}
\label{Conclusions}

This survey has reviewed prompt tuning methods across direct learning and transfer learning. Through analysis of various framework innovations, we observe trends toward improved parameter efficiency, adaptive cross-task learning, and optimized training stability. While progress has been made, challenges remain in computational efficiency, explainability, and model scale dependency. Future directions include robust optimization strategies, scope extension, and interpretability analysis.

\section*{Ethics Statement}
This work does not involve ethical concerns.

\section*{Availability Statement}
Papers and updates related to this survey can be found at https://github.com/ZongqianLi/Prompt-Tuning-Survey.

\bibliography{example_paper}
\bibliographystyle{icml2025}

\newpage
\appendix
\onecolumn

\section{Appendix}

List of papers: 

\noindent \textbf{Direct Prompt Learning: } \\
\textbf{\textit{General: }} \\
Prompt Tuning \citep{PromptTuning},
XPrompt \cite{XPrompt}, 
P Tuning v2 \citep{PTuningv2} \\
\textbf{\textit{Encoder: }} \\
P Tuning \citep{PTuning}, 
Residual Prompt Tuning \citep{ResidualPromptTuning}, 
Prefix Tuning \citep{PrefixTuning} \\
\textbf{\textit{Decomposirion: }} \\
Decomposed Prompt Tuning \citep{DecomposedPromptTuning}, 
DePT \citep{DePT} \\
\textbf{\textit{MoE: }} \\
SMoP \citep{SMoP}, PT-MoE \citep{li2025ptmoeefficientfinetuningframework} \\
\textbf{Transfer Learning: } \\
\textbf{\textit{General: }} \\
SPoT \citep{SPoT}, 
ATTEMPT \citep{ATTEMPT} \\
\textbf{\textit{Encoder: }} \\
TransPrompt \citep{TransPrompt}, 
CTPT \citep{CTPT} \\
\textbf{\textit{Decomposition: }} \\
MPT \citep{MPT} 

This paper was refined with ChatGPT and Claude.

\end{document}